\documentclass[runningheads]{llncs}
\usepackage{graphicx}
\usepackage{amsmath,amssymb} 
\usepackage{color}
\usepackage[width=122mm,left=32mm,paperwidth=186mm,height=193mm,top=32mm,paperheight=257mm]{geometry}

\usepackage[ruled,linesnumbered]{algorithm2e}

\usepackage[numbers,sort&compress]{natbib}

\usepackage{booktabs}

\begin{document}
\pagestyle{headings}
\mainmatter
\def\ECCV18SubNumber{000}  

\title{Deep Co-Training for Semi-Supervised \\Image Recognition} 

\titlerunning{Deep Co-Training for Semi-Supervised Image Recognition}

\authorrunning{Qiao \textit{et al.}}

\author{Siyuan Qiao$^1~$ Wei Shen$^{1,2}~$ Zhishuai Zhang$^1~$ Bo Wang$^3~$ Alan Yuille$^1~$}
\institute{$^1$Johns Hopkins University $^2$Shanghai University $^3$Hikvision Research Institute}

\maketitle

\begin{abstract}
In this paper, we study the problem of semi-supervised image recognition, which is to learn classifiers using both labeled and unlabeled images.
We present Deep Co-Training, a deep learning based method inspired by the Co-Training framework~\cite{CoT}.
The original Co-Training learns two classifiers on two \textit{views} which are data from \textit{different} sources that describe \textit{the same} instances.
To extend this concept to deep learning, Deep Co-Training trains multiple deep neural networks to be the different views and exploits adversarial examples to encourage view difference, in order to prevent the networks from collapsing into each other.
As a result, the co-trained networks provide different and complementary information about the data, which is necessary for the Co-Training framework to achieve good results.
We test our method on SVHN, CIFAR-10/100 and ImageNet datasets, and our method outperforms the previous state-of-the-art methods by a large margin.

\vspace{-0.1in}
\keywords{Co-Training, Deep Networks, Semi-Supervised Learning}
\end{abstract}

\vspace{-0.3in}
\section{Introduction}
Deep neural networks achieve the state-of-art performances in many visual tasks such as
image recognition~\cite{resnet,densenet,alexnet,fewshot,vggnet,GoogleNet,sort,znfnet,gunn} and
object detection~\cite{deeplab,rich,fcnn,fasterrcnn}.
However, training networks requires large-scale labeled datasets~\cite{ILSVRC15,coco} which are usually expensive and difficult to collect.
Given the massive amounts of unlabeled natural images, the idea to use datasets without human annotations becomes very appealing~\cite{ssl}.
In this paper, we study semi-supervised learning with a particular interest in the image recognition problem, the task of which is to use the unlabeled images in addition to the labeled images in order to build better classifiers.

Formally, in the semi-supervised image recognition problem~\cite{stoc_trans,tessl,badgan}, we are provided with a image dataset $\mathcal{D}=\mathcal{S}\cup\mathcal{U}$ where images in $\mathcal{S}$ are labeled and images in $\mathcal{U}$ are not.
The task is to build classifiers on the categories $\mathcal{C}$ in $\mathcal{S}$ using the data in $\mathcal{D}$.
The test data contains only the categories that appear in $\mathcal{S}$.
The problem of learning models on supervised datasets has been extensively studied, and the state-of-the-art methods are deep convolutional networks~\cite{resnet,densenet}.
The core problem is how to use the unlabeled $\mathcal{U}$ to help learning on $\mathcal{S}$.

The method proposed in this paper is inspired by the Co-Training framework~\cite{CoT}, which is an award-winning method for semi-supervised learning in general.
It assumes that each data $x$ in $\mathcal{D}$ has two \textit{views}, \textit{i.e.} $x$ is given as $x=(v_1, v_2)$,
and each view $v_i$ is sufficient for learning an effective model.
For example, the views can have different data sources~\cite{CoT} or different representations~\cite{cotft,cotrans,cotbow}.
Let $\mathcal{X}$ be the distribution that $\mathcal{D}$ is drawn from.
Co-Training assumes that $f_1$ and $f_2$ trained on view $v_1$ and $v_2$ respectively have consistent predictions on $\mathcal{X}$, \textit{i.e.,}
\begin{equation}\label{eq:cot_aspt}
  f(x)=f_1(v_1)=f_2(v_2),~~~\forall x = (v_1, v_2)\sim\mathcal{X} \text{~~~~~(Co-Training Assumption)}
\end{equation}
Based on this assumption, Co-Training proposes a dual-view self-training algorithm:
it first learns a separate classifier for each view on $\mathcal{S}$, and then the predictions of the two classifiers on $\mathcal{U}$ are gradually added to $\mathcal{S}$ to continue the training.
Blum and Mitchell~\cite{CoT} further show that under an additional assumption that the two views of each instance are conditionally independent given the category, Co-Training has PAC-like guarantees on semi-supervised learning.

Given the superior performances of deep neural networks on supervised image recognition, we are interested in extending the Co-Training framework to apply deep learning to semi-supervised image recognition.
A naive implementation is to train two neural networks simultaneously on $\mathcal{D}$ by modeling Eq.~\ref{eq:cot_aspt}.
But this method suffers from a critical drawback: there is no guarantee that the views provided by the two networks give different and complementary information about each data point.
Yet Co-Training is beneficial only if the two views are different, ideally conditionally independent given the category;
after all, there is no point in training two identical networks.
Moreover, the Co-Training assumption encourages the two models to make similar predictions on both $\mathcal{S}$ and $\mathcal{U}$, which can even lead to collapsed neural networks, as we will show by experiments in Section 3.
Therefore, in order to extend the Co-Training framework to take the advantages of deep learning, it is necessary to have a force that pushes networks away to balance the Co-Training assumption that pulls them together.

The force we add to the Co-Training Assumption is \emph{View Difference Constraint} formulated by Eq.~\ref{eq:dif_aspt}, which encourages the networks to be different
\begin{equation}\label{eq:dif_aspt}
  \exists \mathcal{X'}:~f_1(v_1)\neq f_2(v_2),~\forall x = (v_1, v_2)\sim\mathcal{X'} \text{~~~(View Difference Constraint)}
\end{equation}
The challenge is to find a proper and sufficient $\mathcal{X'}$ that is compatible with Eq.~\ref{eq:cot_aspt} (\textit{e.g.} $\mathcal{X'}\cap \mathcal{X}=\varnothing$) and our tasks.
We construct $\mathcal{X}'$ by adversarial examples~\cite{adv}.

In this paper, we present Deep Co-Training (DCT) for semi-supervised image recognition, which extends the Co-Training framework without the drawback discussed above.
Specifically, we model the Co-Training assumption by minimizing the expected Jensen-Shannon divergence between the predictions of the two networks on $\mathcal{U}$.
To avoid the neural networks from collapsing into each other, we impose the view difference constraint by training each network to be resistant to the adversarial examples~\cite{adv,adv2} of the other.
The result of the training is that each network can keep its predictions unaffected on the examples that the other network fails on.
In other words, the two networks provide different and complementary information about the data because they are trained not to make errors at the same time on the adversarial examples for them.
To summarize, the main contribution of DCT is a differentiable modeling that takes into account both the Co-Training assumption and the view difference constraint.
It is a end-to-end solution which minimizes a loss function defined on the dataset $\mathcal{S}$ and $\mathcal{U}$.
Naturally, we extend the dual-view DCT to a scalable multi-view DCT.
We test our method on four datasets, SVHN~\cite{svhn}, CIFAR10/100~\cite{cifar} and ImageNet~\cite{ILSVRC15},
and DCT outperforms the previous state-of-the-arts by a large margin.



\vspace{-0.05in}
\section{Deep Co-Training}
\vspace{-0.04in}
In this section, we present our model of Deep Co-Training (DCT) and naturally extend dual-view DCT to multi-view DCT.
First, we formulate the Co-Training assumption and  the view difference constraint in 2.1 and 2.2, respectively.

\vspace{-0.08in}
\subsection{Co-Training Assumption in DCT}
We start with the dual-view case where we are interested in co-training two deep neural networks for image recognition.
Following the notations in Section 1, we use $\mathcal{S}$ and $\mathcal{U}$ to denote the labeled and the unlabeled dataset.
Let $\mathcal{D}=\mathcal{S}\cup\mathcal{U}$ denote all the provided data.
Let $v_1(x)$ and $v_2(x)$ denote the two views of data $x$.
In this paper, $v_1(x)$ and $v_2(x)$ are convolutional representations of $x$ before the final fully-connected layer $f_i(\cdot)$ that classifies $v_i(x)$ to one of the categories in $\mathcal{S}$.
On the supervised dataset $\mathcal{S}$, we use the standard cross entropy loss
\begin{equation}\label{eq:sup}
  \mathcal{L}_{\text{sup}}(x, y) = H\Big(y, f_1\big(v_1(x)\big)\Big) + H\Big(y, f_2\big(v_2(x)\big)\Big)
\end{equation}
for any data $(x, y)$ in $\mathcal{S}$ where $y$ is the label for $x$ and $H(p, q)$ is the cross entropy between distribution $p$ and $q$.
Training neural networks using only the supervision from $\mathcal{S}$ minimizes the expected loss $\mathbb{E}[\mathcal{L}_{\text{sup}}]$ on all the data of $\mathcal{S}$.

Next, we model the Co-Training assumption.
Co-Training assumes that on the distribution $\mathcal{X}$ where $x$ is drawn from, $f_1(v_1(x))$ and $f_2(v_2(x))$ agree on their predictions.
In other words, we want networks $p_1(x) = f_1(v_1(x))$ and $p_2(x) = f_2(v_2(x))$ to have close predictions on $\mathcal{U}$.
Therefore, we use a natural measure of similarity, the Jensen-Shannon divergence between $p_1(x)$ and $p_2(x)$, \textit{i.e.},
\begin{equation}\label{eq:cot}
  \mathcal{L}_{\text{cot}}(x)
  = H\Big(\dfrac{1}{2}\big(p_1(x) + p_2(x)\big)\Big) - \dfrac{1}{2}\Big(H\big(p_1(x) \big) + H\big(p_2(x)\big)\Big)
\end{equation}
where $x\in\mathcal{U}$ and $H(p)$ is the entropy of $p$.
Training neural networks based on the Co-Training assumption minimizes the expected loss $\mathbb{E}[\mathcal{L}_{\text{cot}}]$ on the unlabeled set $\mathcal{U}$.
As for the labeled set $\mathcal{S}$, minimizing loss $\mathcal{L}_{\text{sup}}$ already encourages them to have close predictions on $\mathcal{S}$ since they are trained with labels; therefore, minimizing $\mathcal{L}_{\text{cot}}$ on $\mathcal{S}$ is unnecessary, and we only implement it on $\mathcal{U}$ (\textit{i.e.} not on $\mathcal{S}$).

\vspace{-0.08in}
\subsection{View Difference Constraint in DCT}
The key condition of Co-Training to be successful is that the two views are different and provide complementary information about each data $x$.
But minimizing Eq.~\ref{eq:sup} and \ref{eq:cot} only encourages the neural networks to output the same predictions on $\mathcal{D}=\mathcal{S}\cup\mathcal{U}$.
Therefore, it is necessary to encourage the networks to be different and complementary.
To achieve this, we create another set of images $\mathcal{D'}$ where $p_1(x)\neq p_2(x)$, $\forall x\in\mathcal{D'}$, which we will generate by adversarial examples~\cite{adv,adv2}.

Since Co-Training assumes that $p_1(x)=p_2(x),~\forall x\in\mathcal{D}$, we know that $\mathcal{D}\cap\mathcal{D'}=\varnothing$.
But $\mathcal{D}$ is all the data we have; therefore, $\mathcal{D'}$ must be built up by a generative method.
On the other hand, suppose that $p_1(x)$ and $p_2(x)$ can achieve very high accuracy on naturally obtained data (\textit{e.g.} $\mathcal{D}$), assuming $p_1(x)\neq p_2(x)$, $\forall x\in\mathcal{D}'$ also implies that $\mathcal{D'}$ should be constructed by a generative method.

We consider a simple form of generative method $g(x)$ which takes data $x$ from $\mathcal{D}$ to build $\mathcal{D'}$, \textit{i.e.} $\mathcal{D}'=\{g(x)~|~x\in\mathcal{D}\}$.
For any $x\in\mathcal{D}$, we want $g(x) - x$ to be small so that $g(x)$ also looks like a natural image.
But when $g(x) - x$ is small, it is very possible that $p_1(g(x)) = p_1(x)$ and $p_2(g(x))=p_2(x)$.
Since Co-Training assumes $p_1(x)=p_2(x)$, $\forall x\in\mathcal{D}$ and we want $p_1(g(x))\neq p_2(g(x))$,
when $p_1(g(x))=p_1(x)$, it follows that $p_2(g(x))\neq p_2(x)$.
These considerations imply that $g(x)$ is an adversarial example~\cite{adv} of $p_2$ that fools the network $p_2$ but not the network $p_1$.
Therefore, in order to prevent the deep networks from collapsing into each other, we propose to train the network $p_1$ (or $p_2$) to be resistant to the adversarial examples $g_2(x)$ of $p_2$ (or $g_1(x)$ of $p_1$) by minimizing the cross entropy between $p_2(x)$ and $p_1(g_2(x))$ (or between $p_1(x)$ and $p_2(g_1(x))$), \textit{i.e.,}
\begin{equation}\label{eq:dif}
  \mathcal{L}_{\text{dif}}(x) = H\Big(p_1(x), p_2\big(g_1(x)\big)\Big) + H\Big(p_2(x), p_1\big(g_2(x)\big)\Big)
\end{equation}

Using artificially created examples in image recognition has be studied before.
They can serve as regularization techniques to smooth outputs~\cite{vat}, or create negative examples to tighten decision boundaries~\cite{badgan,intro_cnn}.
Now, they are used to make networks different.
To summarize the Co-Training with the view difference constraint in a sentence, we want the models to have \emph{the same} predictions on $\mathcal{D}$ but make \emph{different} errors when they are exposed to adversarial attacks.
By minimizing Eq.~\ref{eq:dif} on $\mathcal{D}$, we encourage the models to generate complementary representations, each is resistant to the adversarial examples of the other.

\subsection{Training DCT}
In Deep Co-Training, the objective function is of the form
\begin{equation}\label{eq:los}
  \mathcal{L} = \mathbb{E}_{(x, y)\in\mathcal{S}}\mathcal{L}_{\text{sup}}(x, y) + \lambda_{\text{cot}}\mathbb{E}_{x\in\mathcal{U}}\mathcal{L}_{\text{cot}}(x) +
  \lambda_{\text{dif}}\mathbb{E}_{x\in\mathcal{D}}\mathcal{L}_{\text{dif}}(x)
\end{equation}
which linearly combines Eq.~\ref{eq:sup}, Eq.~\ref{eq:cot} and Eq.~\ref{eq:dif} with hyperparameters $\lambda_{\text{cot}}$ and $\lambda_{\text{dif}}$.
We present one iteration of the training loop in Algorithm~\ref{alg:trn}.
The full training procedure repeats the computations in Algorithm~\ref{alg:trn} for many iterations and epochs using gradient descent with decreasing learning rates.

\begin{algorithm}
\SetKwInOut{Input}{Input}
\SetKwInOut{Output}{Output}
\caption{One Iteration of the Training Loop of Deep Co-Training}\label{alg:trn}
{\bf Data Sampling~} Sample data batch $b_1=(x_{b_1}, y_{b_1})$ for $p_1$ and $b_2=(x_{b_2}, y_{b_2})$ for $p_2$ from $\mathcal{S}$ s.t. $|b_1|=|b_2|=b$.
Sample data batch $b_u=(x_u)$ from $\mathcal{U}$.

{\bf Create Adversarial Examples~} Compute the adversarial examples $g_1(x)$ of $p_1$ for all $x\in x_{b_1}\cup x_u$ and $g_2(x)$ of $p_2$ for all $x\in x_{b_2}\cup x_u$ using FGSM~\cite{adv}.

$\mathcal{L}_{\text{sup}} = \dfrac{1}{b}\Big[\displaystyle\sum_{(x, y)\in b_1}H(y, p_1(x)) + \displaystyle\sum_{(x, y)\in b_2}H(y, p_2(x))\Big]$

$\mathcal{L}_{\text{cot}} = \dfrac{1}{|b_u|}\displaystyle\sum_{x\in b_u} \Big[H\Big(\dfrac{1}{2}\big(p_1(x) + p_2(x)\big)\Big) - \dfrac{1}{2}\Big(H\big(p_1(x) \big) + H\big(p_2(x)\big)\Big)\Big]$

$\mathcal{L}_{\text{dif}} = \dfrac{1}{b + |b_u|}\Big[\displaystyle\sum_{x\in x_1\cup x_u} H(p_1(x), p_2(g_1(x))) + \displaystyle\sum_{x\in x_2\cup x_u} H(p_2(x), p_1(g_2(x)))\Big]$

$\mathcal{L} = \mathcal{L}_{\text{sup}} + \lambda_{\text{cot}}\mathcal{L}_{\text{cot}} +
\lambda_{\text{dif}}\mathcal{L}_{\text{dif}}$

{\bf Update} Compute the gradients with respect to $\mathcal{L}$ by backpropagation and update the parameters of $p_1$ and $p_2$ using gradient descent.
\end{algorithm}

Note that in each iteration of the training loop of DCT, the two neural networks receive different supervised data.
This is to increase the difference between them by providing them with supervised data in different time orders.
Consider that the data of the two networks are provided by two data streams $s$ and $\overline{s}$.
Each data $d$ from $s$ and $\overline{d}$ from $\overline{s}$ are of the form $[d_{s}, d_{u}]$, where $d_{s}$ and $d_{u}$ denote a batch of supervised data and unsupervised data, respectively.
We call $(s, \overline{s})$ \emph{a bundle of data streams} if their $d_{u}$ are the same and the sizes of $d_{s}$ are the same.
Algorithm~\ref{alg:trn} uses a bundle of data streams to provide data to the two networks.
The idea of using bundles of data streams is important for scalable multi-view Deep Co-Training, which we will present in the following subsections.

\subsection{Multi-View DCT}
In the previous subsection, we introduced our model of dual-view Deep Co-Training.
But dual-view is only a special case of multi-view learning, and multi-view co-training has also been studied for other problems~\cite{tritrain,multiview}.
In this subsection, we present a scalable method for multi-view Deep Co-Training.
Here, the scalability means that the hyperparameters $\lambda_{\text{cot}}$ and $\lambda_{\text{dif}}$ in Eq.~\ref{eq:los} that work for dual-view DCT are also suitable for increased numbers of views.
Recall that in the previous subsections, we propose a concept called a bundle of data streams $s=(s, \overline{s})$ which provides data to the two neural networks in the dual-view setting.
Here, we will use multiple data stream bundles to provide data to different views so that the dual-view DCT can be adapted to the multi-view settings.

Specifically, we consider $n$ views $v_i(\cdot)$, $i=1,..,n$ in the multi-view DCT.
We assume that $n$ is a even number for simplicity of presenting the multi-view algorithm.
Next, we build $n/2$ independent data stream bundles $B=\big( (s_1, \overline{s_1}), ..., (s_{n/2}, \overline{s_{n/2}})  \big)$.
Let $B_i(t)$ denote the training data that bundle $B_i$ provides at iteration $t$.
Let $\mathcal{L}(v_i, v_j, B_k(t))$ denote the loss $\mathcal{L}$ in Step $6$ of Algorithm~\ref{alg:trn} when dual training $v_i$ and $v_j$ using data $B_k(t)$.
Then, at each iteration $t$, we consider the training scheme implied by the following loss function
\begin{equation}\label{eq:fake}
  \mathcal{L}_{\text{fake $n$-view}}(t) = \sum_{i=1}^{n/2}\mathcal{L}(v_{2i - 1}, v_{2i}, B_i(t))
\end{equation}
We call this \emph{fake multi-view DCT} because Eq.~\ref{eq:fake} can be considered as $n/2$ independent dual-view DCTs.
Next, we adapt Eq.~\ref{eq:fake} to the \emph{real multi-view DCT}.
In our multi-view DCT, at each iteration $t$, we consider an index list $l$ randomly shuffled  from \{1, 2, .., n\}.
Then, we use the following training loss function
\begin{equation}\label{eq:real}
    \mathcal{L}_{\text{$n$-view}}(t) = \sum_{i=1}^{n/2}\mathcal{L}(v_{l_{2i - 1}}, v_{l_{2i}}, B_i(t))
\end{equation}
Compared with Eq.~\ref{eq:fake}, Eq.~\ref{eq:real} randomly chooses a pair of views to train for each data stream bundle at each iteration.
The benefits of this modeling are multifold.
Firstly, Eq.~\ref{eq:real} is converted from $n/2$ independent dual-view trainings; therefore, the hyperparameters for the dual-view setting are also suitable for multi-view settings.
Thus, we can save our efforts in tuning parameters for different number of views.
Secondly, because of the relationship between Eq.~\ref{eq:fake} and Eq.~\ref{eq:real}, we can directly compare the training dynamics between different number of views.
Thirdly, compared with computing the expected loss on all the possible pairs and data at each iteration, this modeling is also computationally efficient.

\subsection{Implementation Details}
To fairly compare with the previous state-of-the-art methods, we use the training and evaluation framework of Laine and Aila~\cite{tessl}.
We port their implementation to PyTorch for easy multi-GPU support.
Our multi-view implementation will automatically spread the models to different devices for the maximal utilizations.
For SVHN and CIFAR, we use a network architecture similar to \cite{tessl}: we only change their weight normalization and mean-only batch normalization layers~\cite{mobn} to the natively supported batch normalization layers~\cite{batchnorm}.
This change results in performances a little worse than but close to those reported in their paper.
\cite{tessl} thus is the most natural baseline.
For ImageNet, we use a small model ResNet-18~\cite{resnet} for fast experiments.
In the following, we introduce the datasets SVHN, CIFAR and ImageNet, and how we train our models on them.

\paragraph{SVHN~}
The Street View House Numbers (SVHN) dataset~\cite{svhn} contains real-world images of house numbers, each of which is of size $32\times 32$.
The label for each image is the centermost digit.
Therefore, this is a classification problem with $10$ categories.
Following Laine and Aila~\cite{tessl}, we only use $1000$ images out of $73257$ official training images as the supervised part $\mathcal{S}$ to learn the models and the full test set of $26032$ images for testing.
The rest $73257 - 1000$ images are considered as the unsupervised part $\mathcal{U}$.
We train our method with the standard data augmentation, and our method significantly outperforms the previous state-of-the-art methods.
Here, the data augmentation is only the random translation by at most $2$ pixels.
We do not use any other types of data augmentations.

\paragraph{CIFAR~}
CIFAR~\cite{cifar} has two image datasets, CIFAR-10 and CIFAR-100.
Both of them contain color natural images of size $32\times32$,
while CIFAR-10 includes $10$ categories and CIFAR-100 contains $100$ categories.
Both of them have $50000$ images for training and $10000$ images for testing.
Following Laine and Aila~\cite{tessl}, for CIFAR-10, we only use $4000$ images out of $50000$ training images as the supervised part $\mathcal{S}$ and the rest $46000$ images are used as the unsupervised part $\mathcal{U}$. As for CIFAR-100, we use $10000$ images out of $50000$ training images as the supervised part $\mathcal{S}$ and the rest $40000$ images as the unsupervised part $\mathcal{U}$.
We use the full $10000$ test images for evaluation for both CIFAR-10 and CIFAR-100.
We train our methods with the standard data augmentation, which is the combination of random horizontal flip and translation by at most $2$ pixels.

\paragraph{ImageNet}
The ImageNet dataset contains about $1.3$ million natural color images for training and $50000$ images for validation.
The dataset includes $1000$ categories, each of which typically has $1300$ images for training and $50$ for evaluation.
Following the prior work that reported results on ImageNet~\cite{stoc_trans,vae,mean}, we uniformly choose $10\%$ data from $1.3$ million training images as supervised $\mathcal{S}$ and the rest as unsupervised $\mathcal{U}$.
We report the single center crop error rates on the validation set.
We train our models with data augmentation, which includes random resized crop to $224\times 224$ and random horizontal flip.
We do not use other advanced augmentation techniques such as color jittering or PCA lighting~\cite{alexnet}.

For SVHN and CIFAR, following \cite{tessl}, we use a warmup scheme for the hyperparameters $\lambda_{\text{cot}}$ and $\lambda_{\text{dif}}$.
Specifically, we warmup them in the first $80$ epochs such that $\lambda = \lambda_{\text{max}}\cdot \exp(-5 ( 1 -  T / 80) ^ 2)$ when the epoch $T\leq 80$, and $\lambda_{\text{max}}$ after that.
For SVHN and CIFAR, we set $\lambda_{\text{cot,max}}=10$.
For SVHN and CIFAR-10, $\lambda_{\text{dif,max}}=0.5$, and for CIFAR-100 $\lambda_{\text{dif,max}}=1.0$.
For training, we train the networks using stochastic gradient descent with momentum $0.9$ and weight decay $0.0001$.
The total number of training epochs is $600$ and we use a cosine learning rate schedule $lr =  0.05 \times (1.0 + \cos((T - 1) \times \pi / 600))$ at epoch $T$.
The batch size is set to $100$ for SVHN, CIFAR-10 and CIFAR-100.

For ImageNet, we choose a different training scheme.
Before using any data from $\mathcal{U}$, we first train two ResNet-18 individually with different initializations and training sequences on only the labeled data $\mathcal{S}$.
Following ResNet~\cite{resnet}, we train the models using stochastic gradient descent with momentum $0.9$, weight decay $0.0001$ and batch size $256$ for $600$ epochs, the time of which is the same as training $60$ epochs with full supervision.
The learning rate is initialized as $0.1$ and multiplied by $0.1$ at the $301$st epoch.
Then, we take the two pre-trained models to our unsupervised training loops.
This time, we directly set $\lambda$ to the maximum values $\lambda = \lambda_{\text{max}}$ because the previous $600$ epochs have already warmed up the models.
Here, $\lambda_{\text{cot,max}}=1$ and $\lambda_{\text{dif,max}}=0.1$.
In the unsupervised loops, we use a cosine learning rate $lr = 0.005 \times (1.0 + \cos((T - 1) \times \pi / 20))$ and we train the networks for $20$ epochs on both $\mathcal{U}$ and $\mathcal{S}$.
The batch size is set to $128$.

To make the loss $\mathcal{L}$ stable across different training iterations, we require that each data stream provides data batches whose proportions of the supervised data are close to the ratio of the size of $\mathcal{S}$ to the size of $\mathcal{D}$.
To achieve this, we evenly divide the supervised and the unsupervised data to build each data batch in the data streams.
As a result, the difference of the numbers of the supervised images between any two batches is no greater than $1$.

\section{Results}
In this section, we will present the experimental results on four datasets, \textit{i.e.} SVHN~\cite{svhn}, CIFAR-10, CIFAR-100~\cite{cifar} and ImageNet~\cite{ILSVRC15}

\begin{table}
  \setlength{\tabcolsep}{10pt}
  \centering
  \begin{tabular}{lcc}
    \toprule
    Method & SVHN & CIFAR-10 \\
    \midrule
    GAN~\cite{igan} & $8.11\pm1.30$ & $18.63\pm2.32$ \\
    Stochastic Transformations~\cite{stoc_trans} & -- & $11.29\pm0.24$ \\
    $\Pi$ Model~\cite{tessl} & $4.82\pm0.17$ & $12.36\pm0.31$ \\
    Temporal Ensembling~\cite{tessl} & $4.42\pm0.16$ & $12.16\pm0.24$ \\
    Mean Teacher~\cite{mean} & $3.95\pm0.19$ & $12.31\pm0.28$ \\
    Bad GAN~\cite{badgan}  & $4.25\pm0.03$ & $14.41\pm0.30$ \\
    VAT~\cite{vat} & $3.86$ & $10.55$ \\
    \midrule
    Deep Co-Training with 2 Views $~~~~$ & $3.61\pm0.15$ & $9.03\pm0.18$ \\
    Deep Co-Training with 4 Views $~~~~$ & $3.38\pm0.05$ & $8.54\pm0.12$ \\
    Deep Co-Training with 8 Views $~~~~$ & $3.29\pm0.03$ & $8.35\pm0.06$ \\
    \bottomrule
  \end{tabular}
  \vspace{0.05in}
  \caption{Error rates on SVHN (1000 labeled) and CIFAR-10 (4000 labeled) benchmarks.
  Note that we report the averages of the single model error rates without ensembling them for the fairness of comparisons.
  We use architectures that are similar to that of $\Pi$ Model~\cite{tessl}.
  ``--" means that the original papers did not report the corresponding error rates.
  We report means and standard deviations from 5 runs.}
  \label{tab:sc}
\end{table}

\subsection{SVHN and CIFAR-10}
SVHN and CIFAR-10 are the datasets that the previous state-of-the-art methods for semi-supervised image recognition mostly focus on.
Therefore, we first present the performances of our method and show the comparisons with the previous state-of-the-art methods on these two datasets.
Next, we will also provide ablation studies on the two datasets for better understandings of the dynamics and characteristics of dual-view and multi-view Deep Co-Training.

Table~\ref{tab:sc} compares our method Deep Co-Training with the previous state-of-the-art methods on SVHN and CIFAR-10 datasets.
To make sure these methods are fairly compared, we do not ensemble the models of our method even there are multiple well-trained models after the entire training procedure. Instead, we only report the average performances of those models.
Compared with other state-of-the-art methods, Deep Co-Training achieves significant performance improvements when $2$, $4$ or $8$ views are used.
As we will discuss in Section 4, all the methods listed in Table~\ref{tab:sc} require implicit or explicit computations of multiple models, \textit{e.g.} GAN~\cite{igan} has a discriminative and a generative network, Bad GAN~\cite{badgan} adds another encoder network based on GAN, and Mean Teacher~\cite{mean} has an additional EMA model.
Therefore, the dual-view Deep Co-Training does not require more computations in terms of the total number of the networks.

Another trend we observe is that although $4$-view DCT gives significant improvements over $2$-view DCT, we do not see similar improvements when we increase the number of the views to $8$.
For this observation, we speculate that this is because compared with $2$-views, $4$-views can use the majority vote rule when we encourage them to have close predictions on $\mathcal{U}$.
When we increase the number of views to $8$, although it is expected to perform better, the advantages over $4$-views are not that strong compared with that of $4$-views over $2$-views.
But $8$-view DCT converges faster than $4$-view DCT, which is even faster than dual-view DCT.
The training dynamics of DCT with different numbers of views will be presented in the later subsections.
We first provide our results on CIFAR-100 and ImageNet datasets in the next subsection.

\begin{table}
  \setlength{\tabcolsep}{4pt}
  \centering
  \begin{tabular}{lcc}
    \toprule
    Method & CIFAR-100 & CIFAR-100+ \\
    \midrule
    $\Pi$ Model~\cite{tessl} & $43.43\pm0.54$ & $39.19\pm0.36$\\
    Temporal Ensembling~\cite{tessl} & -- & $38.65\pm0.51$ \\
    \midrule
    Dual-View Deep Co-Training & $\mathbf{38.77\pm0.28}$ & $\mathbf{34.63\pm0.14}$ \\
    \bottomrule
  \end{tabular}
  \vspace{0.05in}
  \caption{Error rates on CIFAR-100 with $10000$ images labeled.
  Note that other methods listed in Table~\ref{tab:sc} have not published results on CIFAR-100.
  The performances of our method are the averages of single model error rates of the networks without ensembling them for the fairness of comparisons.
  We use architectures that are similar to that of $\Pi$ Model~\cite{tessl}.
  ``--" means that the original papers did not report the corresponding error rates.
  CIFAR-100+ and CIFAR-100 indicate that the models are trained with and without data augmentation, respectively.
  Our results are reported from $5$ runs.}
  \label{tab:c100}
\end{table}

\subsection{CIFAR-100 and ImageNet}

Compared with SVHN and CIFAR-10, CIFAR-100 and ImageNet are considered harder benchmarks~\cite{tessl} for the semi-supervised image recognition problem because their numbers of categories are 100 and 1000, respectively,  greater than 10 categories in SVHN and CIFAR-10.
Here, we provide our results on these two datasets.
Table~\ref{tab:c100} compares our method with the previous state-of-the-art methods that report the performances on CIFAR-100 dataset, \textit{i.e.} $\Pi$ Model and Temporal Ensembling~\cite{tessl}.
Dual-view Deep Co-Training even without data augmentation achieves similar performances with the previous state-of-the-arts that use data augmentation.
When our method also uses data augmentation, the error rate drops significantly from $38.65$ to $34.63$.
These results demonstrate the effectiveness of the proposed Deep Co-Training when the number of categories and the difficulty of the datasets increase.

\begin{table}
  \setlength{\tabcolsep}{6pt}
  \centering
  \begin{tabular}{llccc}
    \toprule
    Method & Architecture & \# Param & Top-1 & Top-5 \\
    \midrule
    Stochastic Transformations~\cite{stoc_trans} & AlexNet & 61.1M & -- & 39.84 \\
    VAE~\cite{vae} with $10\%$ Supervised & Customized & 30.6M & 51.59 & 35.24 \\
    Mean Teacher~\cite{mean} & ResNet-18 & 11.6M & 49.07 & 23.59 \\
    \midrule
    $100\%$ Supervised & ResNet-18 & 11.6M & 30.43 & 10.76 \\
    $10\%$ Supervised & ResNet-18 & 11.6M & 52.23 & 27.54 \\
    Dual-View Deep Co-Training & ResNet-18 & 11.6M & \bf 46.50 & \bf 22.73 \\
    \bottomrule
  \end{tabular}
  \vspace{0.05in}
  \caption{Error rates on the validation set of ImageNet benchmark with $10\%$ images labeled.
  The image size of our method in training and testing is $224\times 224$.}
  \label{tab:imagenet}
\end{table}

\begin{figure}
  \includegraphics[width=\linewidth]{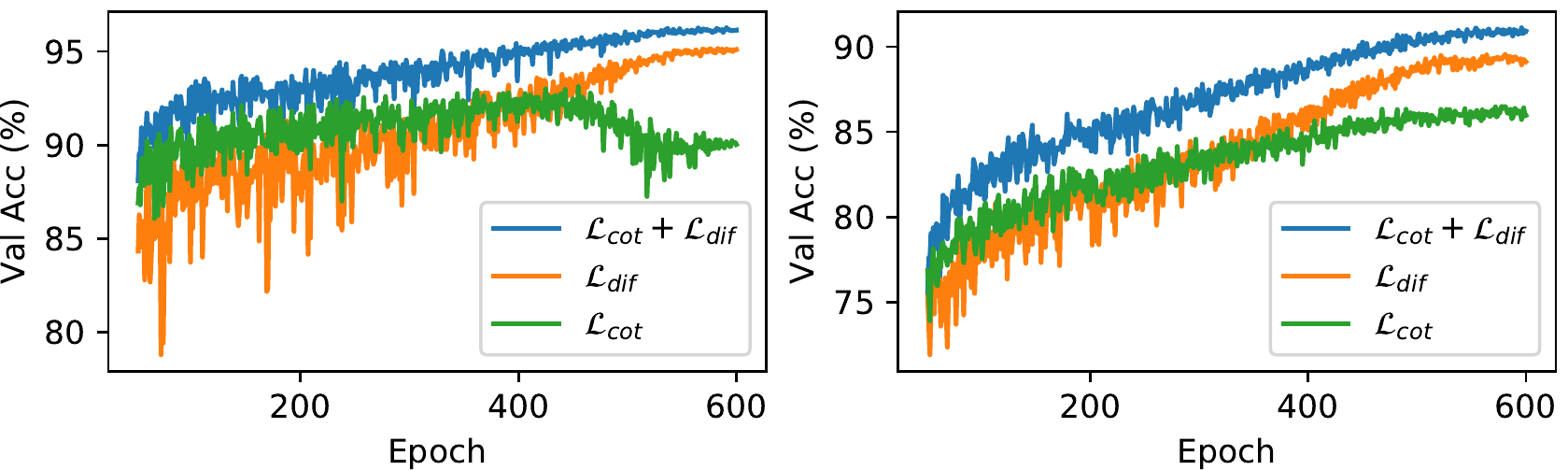}
  \caption{Ablation study on $\mathcal{L}_{\text{cot}}$ and $\mathcal{L}_{\text{dif}}$.
  The left plot is the training dynamics of dual-view Deep Co-Training on SVHN dataset,
  and the right is on CIFAR-10 dataset.
  ``$\lambda_{\text{cot}}$", ``$\lambda_{\text{dif}}$" represent the loss functions are used alone while ``$\lambda_{\text{cot}}+\lambda_{\text{dif}}$" correspond to the weighted sum loss used in Deep Co-Training.
  In all the cases, $\mathcal{L}_{\text{sup}}$ is used.
  }
  \label{fig:cot_dif}
\end{figure}

Next, we show our results on ImageNet with $1000$ categories and $10\%$ labeled in Table~\ref{tab:imagenet}.
Our method has better performances than the supervised-only but is still behind the accuracy when $100\%$ supervision is used.
When compared with the previous state-of-the-art methods, however, DCT shows significant improvements on both the Top-1 and Top-5 error rates.
Here, the performances of \cite{stoc_trans} and \cite{vae} are quoted from their papers, and the performance of Mean Teacher~\cite{mean} with ResNet-18~\cite{resnet} is from running their official implementation on GitHub.
When using the same architecture, DCT outperforms Mean Teacher by $\sim 2.6\%$ for Top-1 error rate, and $\sim 0.9\%$ for Top-5 error rate.
Compared with \cite{stoc_trans} and \cite{vae} that use networks with more parameters and larger input size $256\times256$, Deep Co-Training also achieves lower error rates.

\subsection{Ablation Study}
In this subsection, we will provide several ablation studies for better understandings of our proposed Deep Co-Training method.

\vspace{-0.05in}
\subsubsection{On $\mathcal{L}_{\text{cot}}$ and $\mathcal{L}_{\text{dif}}$}
Recall that the loss function used in Deep Co-Training has three parts, the supervision loss $\mathcal{L}_{\text{sup}}$, the co-training loss $\mathcal{L}_{\text{cot}}$ and the view difference constraint $\mathcal{L}_{\text{dif}}$.
Both $\mathcal{L}_{\text{cot}}$ and $\mathcal{L}_{\text{dif}}$ provide certain amounts of supervisions based on different assumptions.
Therefore, it is of interest to study the changes when the loss function $\mathcal{L}_{\text{cot}}$ and $\mathcal{L}_{\text{dif}}$ are used alone in addition to $\mathcal{L}_{\text{sup}}$ in $\mathcal{L}$.
Fig.~\ref{fig:cot_dif} shows the plots of the training dynamics of Deep Co-Training when different loss functions are used on SVHN and CIFAR-10 dataset.
In both plots, the blue lines represent the loss function that we use in practice in training DCT, the green lines represent only the co-training loss $\mathcal{L}_{\text{cot}}$ and $\mathcal{L}_{\text{sup}}$ are applied, and the orange lines represent only the the view difference constraint $\mathcal{L}_{\text{dif}}$ and $\mathcal{L}_{\text{sup}}$ are used.
From Fig.~\ref{fig:cot_dif}, we can see that the Co-Training assumption ($\mathcal{L}_{\text{cot}}$) performs better at the beginning, but soon is overtaken by $\mathcal{L}_{\text{dif}}$.
$\mathcal{L}_{\text{cot}}$ even falls into an extreme case in the SVHN dataset where its validation accuracy drops suddenly around the $400$-th epoch.
However, the loss $\mathcal{L}_{\text{sup}}$ and  $\mathcal{L}_{\text{cot}}$ around that epoch look smooth and normal.
For this phenomenon, we speculate that this is because the networks have collapsed into each other, which motivates us to investigate the dynamics of loss $\mathcal{L}_{\text{dif}}$.
If our speculation is correct, there will also be abnormalities in loss $\mathcal{L}_{\text{dif}}$ around that epoch, which indeed we show in the next subsection.
Moreover, this also supports our argument at the beginning of the paper that a force to push models away is necessary for co-training multiple neural networks for semi-supervised learning.

\begin{figure}[t]
  \includegraphics[width=\linewidth]{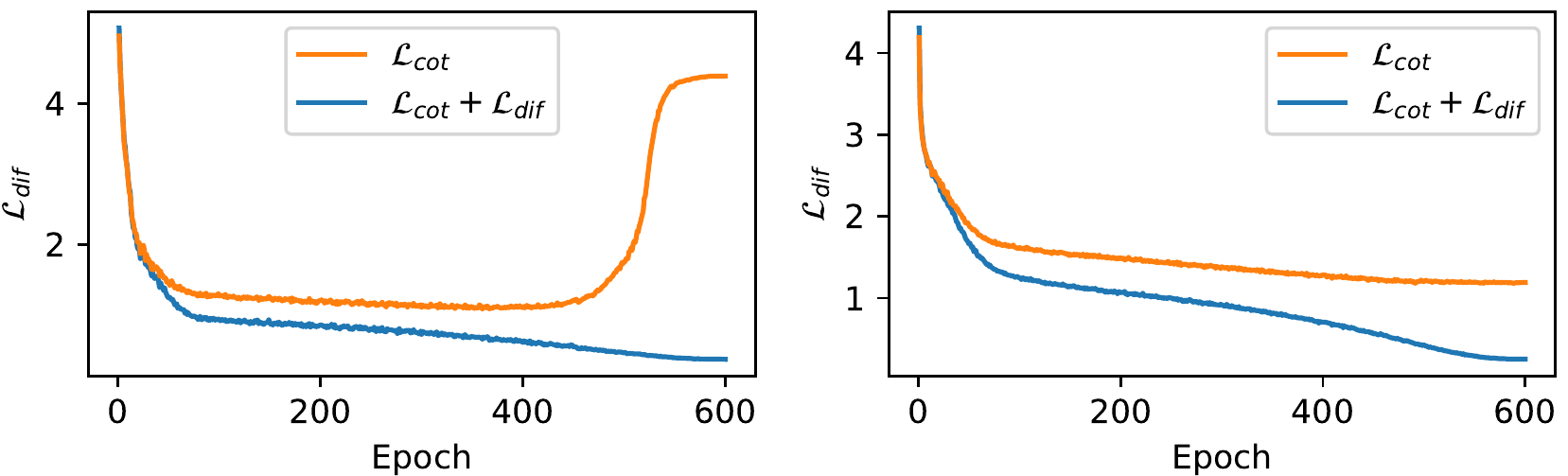}
  \caption{Ablation study on the view difference.
  The left plot is $\mathcal{L}_{\text{dif}}$ on SVHN dataset, and the right plot shows $\mathcal{L}_{\text{dif}}$ on CIFAR-10.
  Without minimizing $\mathcal{L}_{\text{dif}}$, $\mathcal{L}_{\text{dif}}$ is usually big in ``$\mathcal{L}_{\text{cot}}$", indicating that the two models are making similar errors.
  In the SVHN dataset, the two models start to collapse into each other after around the $400$-th epoch because we observe a sudden increase of $\mathcal{L}_{\text{dif}}$.
  This corresponds to the sudden drop in the left plot of Fig.~\ref{fig:cot_dif}, which shows the relation between view difference and accuracy.
  }
  \label{fig:dif}
\end{figure}
\subsubsection{On the view difference}
This is a sanity check on whether in dual-view training, two models tend to collapse into each other when we only model the Co-Training assumption, and if $\mathcal{L}_{\text{dif}}$ can push them away during training.
To study this, we plot $\mathcal{L}_{\text{dif}}$ when it is minimized as in the Deep Co-Training and when it is not minimized, \textit{i.e.} $\lambda_{\text{dif}}=0$.
Fig.~\ref{fig:dif} shows the plots of $\mathcal{L}_{\text{dif}}$ for SVHN dataset and CIFAR dataset, which correspond to the validation accuracies shown in Fig.~\ref{fig:cot_dif}.
It is clear that when $\mathcal{L}_{\text{dif}}$ is not minimized as in the ``$\mathcal{L}_{\text{cot}}$" case,
$\mathcal{L}_{\text{dif}}$ is far greater than $0$, indicating that each model is vulnerable to the adversarial examples of the other.
Like the extreme case we observe in Fig.~\ref{fig:cot_dif} for SVHN dataset (left) around the $400$-th epoch, we also see a sudden increase of $\mathcal{L}_{\text{dif}}$ here in Fig.~\ref{fig:dif} for SVHN at the similar epochs.
This means that every adversarial example of one model fools the other model, \textit{i.e.} they collapse into each other.
The collapse directly causes a significant drop of the validation accuracy in the left of Fig.~\ref{fig:cot_dif}.
These experimental results demonstrate the positive correlation between the view difference and the validation error.
It also shows that the models in the dual-view training tend to collapse into each other when no force is applied to push them away.
Finally, these results also support the effectiveness of our proposed $\mathcal{L}_{\text{dif}}$ as a loss function to increase the difference between the models.

\subsubsection{On the number of views}
\begin{figure}[t]
  \includegraphics[width=\linewidth]{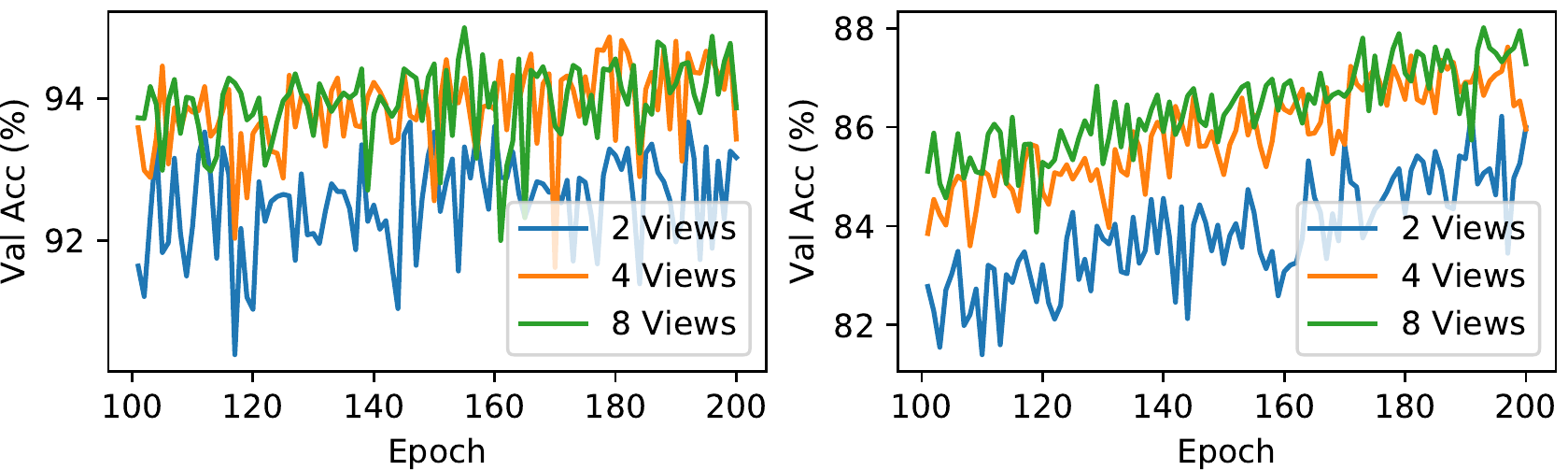}
  \caption{Training dynamics of Deep Co-Training with different numbers of views on SVHN dataset (left) and CIFAR-10 (right).
  The plots focus on the epochs from 100 to 200 where the differences are clearest.
  We observe a faster convergence speed when the number of views increases, but the improvements become smaller when the numbers of views increase from 4 to 8 compared with that from 2 to 4.
  }
  \label{fig:view}
\end{figure}

We have provided the performances of Deep Co-Training with different numbers of views for SVHN and CIFAR-10 datasets in Table~\ref{tab:sc}, where we show that increasing the number of the views from 2 to 4 improves the performances of each individual model.
But we also observe that the improvement becomes smaller when we further increase the number of views to 8.
In Fig.~\ref{fig:view}, we show the training dynamics of Deep Co-Training when different numbers of views are trained simultaneously.

As shown in Fig.~\ref{fig:view}, we observe a faster convergence speed when we increase the number of views to train simultaneously.
We focus on the epochs from 100 to 200 where the differences between different numbers of views are clearest.
The performances of different views are directly comparable because of the scalability of the proposed multi-view Deep Co-Training.
Like the improvements of $8$ views over $4$ views on the final validation accuracy, the improvements of the convergence speed also decrease compared with that of $4$ views over $2$ views.
The experimental results in Table~\ref{tab:sc} and here suggest that $4$-view DCT achieves a good balance between the performance and the computation efficiency.

\section{Discussions}
In this section, we discuss the relationship between Deep Co-Training and the previous methods.
We also present perspectives alternative to the Co-Training framework for discussing Deep Co-Training.

\subsection{Related Work}
Deep Co-Training is also inspired by the recent advances in semi-supervised image recognition techniques~\cite{stoc_trans,tessl,vat,PseudoEnsembles,ladder1} which train deep neural networks $f(\cdot)$ to be resistant to noises $\epsilon(z)$, \textit{i.e.} $f(x) = f(x + \epsilon(z))$.
We notice that their computations in one iteration require double feedforwardings and backpropagations, one for $f(x)$ and one for $f(x+\epsilon(z))$.
We ask the question: what would happen if we train two individual models as doing so requires the same amount of computations?
We soon realized that training two models and encouraging them to have close predictions is related to the Co-Training framework~\cite{CoT}, which has good theoretical results, provided that the two models are conditional independent given the category.
However, training models with only the Co-Training assumption is not sufficient for getting good performances because the models tend to collapse into each other, which is against the view difference between different models which is necessary for the Co-Training framework.

As stated in 2.2, we need a generative method to generate images on which two models predict differently.
Generative Adversarial Networks (GANs)~\cite{badgan,igan,wgan} are popular generative models for vision problems, and have also been used for semi-supervised image recognition.
A problem of GANs is that they will introduce new networks to the Co-Training framework for generating images, which also need to be learned.
Compared with GANs, Introspective Generative Models~\cite{intro_cnn,intro} can generate images from discriminative models in a lightweight manner, which bears some similarities with the adversarial examples~\cite{adv}.
The generative methods that use discriminative models also include DeepDream~\cite{deepdream}, Neural Artistic Style~\cite{artistic}, \textit{etc}.
We use adversarial examples in our Deep Co-Training for its natural applicability to avoiding models from collapsing into each other by training each model with the adversarial examples of the others.

Before the work discussed above, semi-supervised learning in general has already been widely studied.
For example, the mutual-exclusivity loss used in \cite{stoc_trans} and the entropy minimization used in \cite{vat} resemble soft implementations of the self-training technique~\cite{selftraining1,selftraining2}, one of the earliest approaches for semi-supervised classification tasks.
\cite{ssl} provides a good survey for the semi-supervised learning methods in general.

\subsection{Alternative Perspectives}
In this subsection, we discuss the proposed Deep Co-Training method from several perspectives alternative to the Co-Training framework.

\subsubsection{Model Ensemble}
Ensembling multiple independently trained models to get a more accurate and stable classifier is a widely used technique to achieve higher performances~\cite{bagging}.
This is also applicable to deep neural networks~\cite{ensemble1,ensemble2}.
In other words, this suggests that when multiple networks with the same architecture are initialized differently and trained using data sequences in different time orders, they can achieve similar performances but in a complementary way~\cite{twoculture}.
In multi-view Deep Co-Training, we also train multiple models in parallel, but not independently, and our evaluation is done by taking one of them as the final classifier instead of averaging their predicted probabilities.
Deep Co-Training in effect is searching for an initialization-free and data-order-free solution.

\subsubsection{Multi-Agent Learning}
After the literature review of the most recent semi-supervised learning methods for image recognition, we find that almost all of them are within the multi-agent learning framework~\cite{malearn}.
To name a few, GAN-based methods at least have a discriminative network and a generative network.
Bad GAN~\cite{badgan} adds an encoder network based on GAN.
The agents in GANs are interacting in an adversarial way.
As we stated in Section 4.1, the methods that train deep networks to be resistant to noises also have the interacting behaviors as what two individual models would have, \textit{i.e.} double feedforwardings and backpropagations.
The agents in these methods are interacting in a cooperative way.
Deep Co-Training explicitly models the cooperative multi-agent learning, which trains multiple agents from the supervised data and cooperative interactions between different agents.
In the multi-agent learning framework, $\mathcal{L}_{\text{dif}}$ can be understood as learning from the errors of the others, and the loss function Eq.~\ref{eq:real} resembles the simulation of interactions within a crowd of agents.

\subsubsection{Knowledge Distillation}
One characteristic of Deep Co-Training is that the models not only learn from the supervised data, but also learn from the predictions of the other models.
This is reminiscent to knowledge distillation~\cite{distill} where student models learn from teacher models instead of the supervisions from the datasets.
In Deep Co-Training, all the models are students and learn from not only the predictions of the other student models but also the errors they make.

\section{Conclusion}
In this paper, we present Deep Co-Training, a method for semi-supervised image recognition.
It extends the Co-Training framework, which assumes that the data has two complementary views, based on which two effective classifiers
can be built and are assumed to have close predictions on the unlabeled images.
Motivated by the recent successes of deep neural networks in supervised image recognition, we extend the Co-Training framework to apply deep networks to the task of semi-supervised image recognition.
In our experiments, we notice that the models are easy to collapse into each other, which violates the requirement of the view difference in the Co-Training framework.
To prevent the models from collapsing, we use adversarial examples as the generative method to generate data on which the views have different predictions.
The experiments show that this additional force that pushes models away is helpful for training and improves accuracies significantly compared with the Co-Training-only modeling.

Since Co-Training is a special case of multi-view learning, we also naturally extend the dual-view DCT to a scalable multi-view Deep Co-Training method where the hyperparameters for two views are also suitable for increased numbers of views.
We test our proposed Deep Co-Training on the SVHN and CIFAR-10 datasets which are the benchmarks that the previous state-of-the-art methods are tested on.
Our method outperforms them by a large margin with $2$, $4$ and $8$ views.
We further provide our results on harder benchmark CIFAR-100 and ImageNet datasets that most of the previous methods have not reported their performances on.
The experimental results demonstrate the effectiveness of our method for the problem of semi-supervised image recognition.
We also provide alternative perspectives for discussing the proposed Deep Co-Training method, including model ensemble, multi-agent learning and knowledge distillation.


\small
\bibliographystyle{splncs}
\bibliography{egbib}
\end{document}